\documentclass[letterpaper]{article} 
\usepackage{aaai24}  
\usepackage{times}  
\usepackage{helvet}  
\usepackage{courier}  
\usepackage[hyphens]{url}  
\usepackage{graphicx} 
\usepackage{natbib}  
\usepackage{caption} 
\usepackage{algorithm}
\usepackage{algorithmic}
\usepackage{amsmath}
\usepackage{amssymb}
\usepackage{booktabs}
\usepackage{xcolor}
\usepackage{multirow}
\usepackage{multicol}

\usepackage{newfloat}
\usepackage{listings}
\DeclareCaptionStyle{ruled}{labelfont=normalfont,labelsep=colon,strut=off} 
\floatstyle{ruled}
\newfloat{listing}{tb}{lst}{}
\floatname{listing}{Listing}
\DeclareMathOperator*{\argtopk}{arg\,max_{top(k)}}

\title{Progressive Feature Self-Reinforcement for \\Weakly Supervised Semantic Segmentation}
\author{
    Jingxuan He\textsuperscript{\rm 1},
    Lechao Cheng\textsuperscript{\rm 1}$\thanks{corresponding author.}$,
    Chaowei Fang\textsuperscript{\rm 3},
    Zunlei Feng\textsuperscript{\rm 2},
    Tingting Mu\textsuperscript{\rm 4},
    Mingli Song\textsuperscript{\rm 2}
}
\affiliations {
    \textsuperscript{\rm 1}Zhejiang Lab, 
    \textsuperscript{\rm 2}Zhejiang University, 
    \textsuperscript{\rm 3}Xidian University, 
    \textsuperscript{\rm 4}University of Manchester\\
    22021025@zju.edu.cn, chenglc@zhejianglab.com, chaoweifang@outlook.com, zunleifeng@zju.edu.cn, tingting.mu@manchester.ac.uk, brooksong@zju.edu.cn
}

\begin{document}

\maketitle

\begin{abstract}
Compared to conventional semantic segmentation with pixel-level supervision, weakly supervised semantic segmentation (WSSS) with image-level labels poses the challenge that it always focuses on the most discriminative regions, resulting in a disparity between fully supervised conditions.
A typical manifestation is the diminished precision on the object boundaries, leading to a deteriorated accuracy of WSSS. 
To alleviate this issue, we propose to adaptively partition the image content into certain regions (e.g., confident foreground and background) and uncertain regions (e.g., object boundaries and misclassified categories) for separate processing. 
For uncertain cues, we propose an adaptive masking strategy and seek to recover the local information with self-distilled knowledge. 
We further assume that the unmasked confident regions should be robust enough to preserve the global semantics. Building upon this, we introduce a complementary self-enhancement method that constrains the semantic consistency between these confident regions and an augmented image with the same class labels. 
Extensive experiments conducted on PASCAL VOC 2012 and MS COCO 2014 demonstrate that our proposed single-stage approach for WSSS not only outperforms state-of-the-art benchmarks remarkably but also surpasses multi-stage methodologies that trade complexity for accuracy. The code can be found at \textit{https://github.com/Jessie459/feature-self-reinforcement}. 
\end{abstract}

\section{Introduction}

Weakly supervised semantic segmentation (WSSS) reduces the cost of annotating ``strong" pixel-level labels by using ``weak" labels such as bounding boxes~\cite{dai2015boxsup,song2019box}, scribbles~\cite{lin2016scribblesup,vernaza2017learning}, points~\cite{bearman2016s, su2022sasformer} and image-level class labels~\cite{araslanov2020single, ru2022learning, wu2023masked, ru2023token}. 
Among these, image-level class labels are the most affordable, but challenging to exploit. 
A commonly used WSSS approach based on image-level class labels typically includes the following steps: (1) to train a neural network for image classification; (2) to  use the network to generate class activation maps (CAMs)~\cite{zhou2016learning} as seed regions; (3) to refine the CAMs to pseudo segmentation labels that will be used as the ground truth for supervising a segmentation network.
These steps can either be implemented as separate stages or as a single collaborative stage, and single-stage frameworks are usually more efficient as they streamline the training pipeline. 
In general, high-quality pseudo labels lead to superior semantic segmentation performance.
In this work, we focus on developing an effective single-stage approach to generate more accurate pseudo labels from image-level class labels.

\begin{figure}[t]
    \centering
    \includegraphics[width=1.0\linewidth]{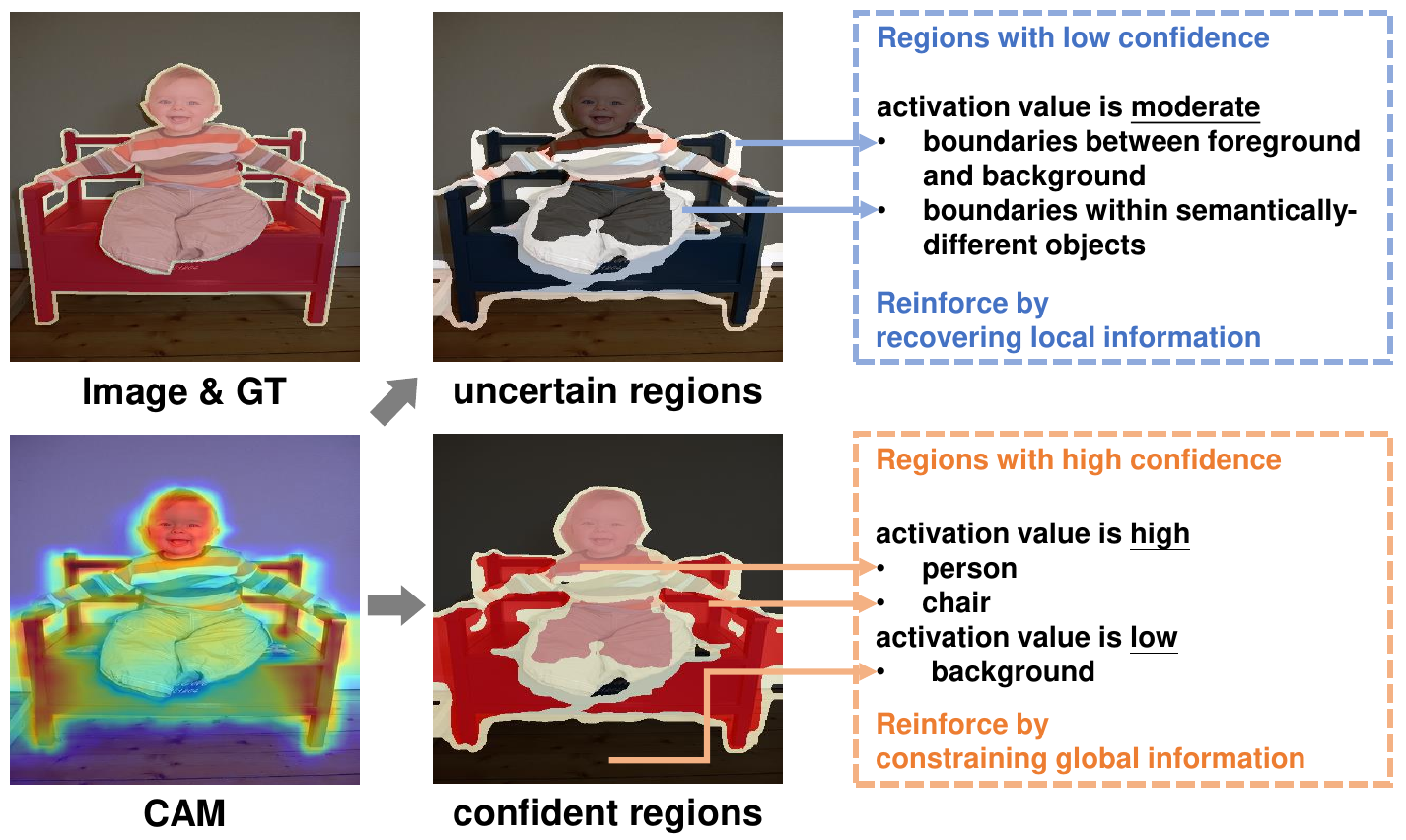}
    \caption{Our main idea. The flawed CAM only identifies discriminative regions. To solve this, we propose to partition the image into uncertain regions (e.g., object boundaries) and confident regions (e.g., the main body of an object) and reinforce features of these regions in a complementary way.}
    \label{fig:teaser}
\end{figure}

Unfortunately, CAMs are essentially flawed because they are intended for classification, i.e., they strive to identify the most discriminative regions of an object aiming at improved classification accuracy. 
To tackle this, one can improve the initial seeds~\cite{lee2019ficklenet,wang2020self} or refine pseudo labels~\cite{ahn2019weakly,chen2020weakly}, by 
expanding activations or labels to semantically consistent pixels in the neighborhood.
Recent studies have found that the restricted receptive field of convolution negatively affects the recognition of integral objects~\cite{ru2022learning,ru2023token} and  use vision transformer~\cite{dosovitskiy2020image} to model the global relationships for improvement. 
But this does not resolve the issue of CAM seeds or pseudo labels, and  we still  observe empirically high uncertainty in  (1) boundary regions between foreground objects and background, and (2) misclassified regions within multiple semantically-different objects. 
In the example of Figure~\ref{fig:teaser}, the generated CAM is uncertain about the two arms of the \textit{person} on the \textit{chair}, also the boundary between the foreground (\textit{person} and \textit{chair}) and the background is unclear. 
These uncertain regions are easily confused by obscure semantic clues due to the absence of pixel-level supervision. 

Our goal is to clarify the visual semantics of uncertain regions mentioned above. 
We emphasize that the local visual patterns should be explicitly modeled and captured. As can be seen from Figure~\ref{fig:teaser}, \textit{head} and \textit{upper thighs} are well recognized, while the recognition of \textit{arms} and \textit{lower legs} needs improvement. 
A better understanding of that \textit{arms and lower legs surely belong to a person} should be established using local visual context. 
Although some methods can deal with noisy object boundaries by employing off-the-shelf saliency detection models for rich object contours~\cite{lee2021railroad,li2022towards},  they overlook uncertain regions caused by low confidence within objects. 
Alternatively, it has been proposed to attain the training objective using knowledge gathered from the past training iterations, i.e., self-distillation~\cite{caron2021emerging}. 
Encouraged by the success of self-distillation, we discard saliency detection models, but  take advantage of  the strategy of self-distillation in our model training.

To this end,  to explore and strengthen semantics over uncertain regions, we propose a progressive self-reinforcement method. 
To distinguish uncertain regions from confident ones, we define those with intermediate CAM scores as uncertain regions, since a very low/high score strongly indicates the background/foreground. 
Specifically, we propose to mask uncertain features (equivalent to image patch tokens) and learn to recover the original information with the help of an online momentum teacher. 
This masking strategy aligns with a state-of-the-art pre-training paradigm called masked image modeling (MIM) that brings locality inductive bias to the model~\cite{xie2023revealing}. 
We upgrade its random masking strategy with semantic uncertainty so that the network can focus on uncertain features controlled by the masking ratio. 
This design is beneficial to facilitate features in both object boundaries and misclassified regions. 
Assuming that confident features should be robust enough to present global semantics, we also introduce a complementary method that constrains semantic consistency between two augmented views with the same class labels. 
Our proposal can be seamlessly integrated into a vision transformer based single-stage WSSS framework. 
We summarize our main contributions as follows :
\begin{itemize}
    \item We propose a novel WSSS approach, \emph{progressive feature self-reinforcement}, to effectively enhance the semantics of uncertain regions. The investigation of uncertain regions, including both object boundaries and misclassified categories, significantly improves WSSS performance. 
    \item We design an adaptive masking strategy to identify uncertain regions. Unlike most previous works that adopt additional saliency detection models, we locate uncertain regions with the guidance of semantic-aware CAMs.
    \item Exhaustive experiments on PASCAL VOC 2012~\cite{everingham2010pascal} and MS COCO 2014~\cite{lin2014microsoft} show that our method outperforms SOTA single-stage competitors, even better than existing sophisticated multi-stage methods. 
\end{itemize}

\section{Related Work}

\begin{figure*}[t]
    \centering
    \includegraphics[width=0.9\linewidth]{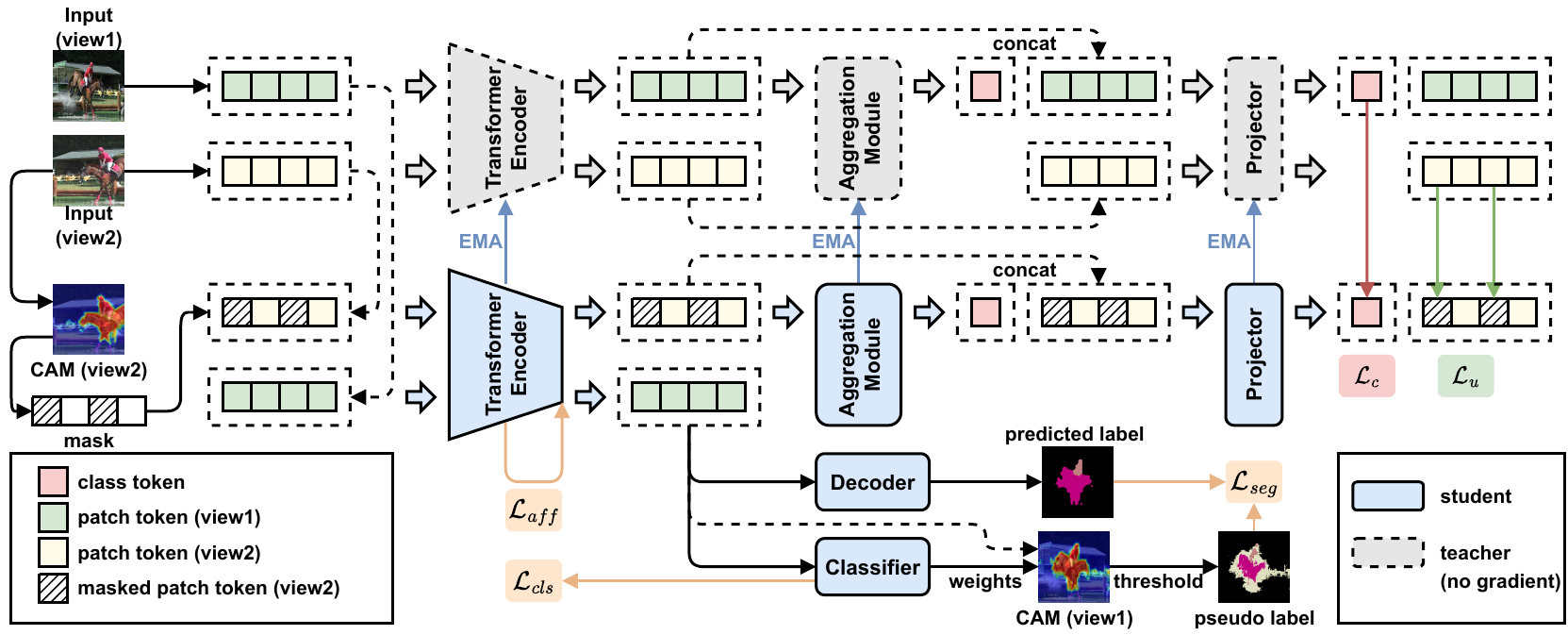}
    \caption{Overview of our framework. For the student pipeline, we forward one view through the encoder, and the encoded patch tokens are fed into the classifier for classification and the decoder for semantic segmentation, separately. The other view is masked and sequentially forwarded through the encoder, the aggregation module, and the projector. For the teacher pipeline, both views are propagated through the encoder, the aggregation module, and the projector. The teacher network requires no gradient and is an exponential moving average (EMA) of the student network.}
    \label{fig:overview}
\end{figure*}

\subsection{Weakly Supervised Semantic Segmentation}
\textbf{Multi-stage WSSS methods} adopt a classification model to generate CAMs as pseudo labels, then train a segmentation model for evaluating the final performance. 
To overcome the commonly acknowledged weakness that CAMs can only focus on discriminative regions, several works aim at improving the training dynamic by erasing and seeking~\cite{hou2018self} or adversarial learning~\cite{yoon2022adversarial}. 
Some recent approaches also adopt  vision transformer ~\cite{dosovitskiy2020image} for the WSSS task, considering its favorable long-range modeling capability. TS-CAM~\cite{gao2021ts} proposes to couple class-agnostic attention maps with semantic-aware patch tokens to promote object localization. MCTformer~\cite{xu2022multi} introduces multiple class tokens so that class-specific attention maps can be generated. Other approaches incorporate extra data into training or post-processing, e.g., saliency maps~\cite{lee2021railroad} or contrastive language-image pre-training (CLIP) models~\cite{lin2023clip}. Our solution aims at improving pseudo labels as well, but it is integrated into a single-stage framework for simplicity, and it requires neither extra data nor off-the-shelf saliency detection models. 

\textbf{Single-stage WSSS methods} treat multiple stages such as classification, pseudo label refinement, segmentation as a whole and perform joint training. 1Stage~\cite{araslanov2020single} achieves comparable performance with dominant multi-stage approaches by ensuring local consistency, semantic fidelity and mask completeness. AFA~\cite{ru2022learning} explores the intrinsic architecture of ViT and derives reliable semantic affinity from multi-head self-attention for pseudo label refinement. ToCo~\cite{ru2023token} tackles the issue of over-smoothing observed in ViT by supervising the final patch tokens with intermediate knowledge. Despite the simplified and streamlined training procedure, single-stage methods often suffer from inferior performance compared with multi-stage ones. In this work, we achieve superior semantic segmentation results using a single-stage framework by discovering and reinforcing underlying semantic layouts.

\subsection{Self-Distillation}
Self-distillation associates self-supervised learning~\cite{he2020momentum} with {knowledge distillation}~\cite{hinton2015distilling}, where knowledge is transferred and learned without resorting to any labels. It is primarily designed to compress large networks, and is hoping to promote performance on downstream tasks via mimicking the output of a frozen teacher~\cite{noroozi2018boosting, zhang2023generalization, wang2023improving}. Recently, some approaches \cite{caron2021emerging,zhou2021ibot} build the teacher dynamically during training, where the teacher adopts the same architecture as that of the student and is updated with the knowledge of past iterations. The resulting framework simplifies the training process and achieves compelling results compared with other self-training frameworks. 
This motivates us to adapt the core idea of self-distillation to the WSSS task for the purpose of rectifying inaccurate object boundaries as well as improving discriminative object features. 

\section{Methodology}

\subsection{A Single-Stage Framework for WSSS}
The proposed single-stage framework for WSSS is illustrated in Figure~\ref{fig:overview}. 
We use an encoder-decoder architecture to accomplish semantic segmentation with image-level supervision. 
The encoder is a vision transformer supervised by image-level class labels. 
We adopt patch token contrast (PTC)~\cite{ru2023token} for affinity learning as it is crucial to constrain affinities between patch tokens of the last layer against over-smoothing~\cite{gong2021vision}. 
As for semantic segmentation, we borrow a lightweight convolutional decoder from DeepLab~\cite{chen2017deeplab}, which is supervised by   pseudo segmentation labels that are generated from CAMs. 
An aggregation module is used to summarize  patch tokens into one class token and an MLP-based projector to transform all tokens into an appropriate feature space for feature learning. 
To improve model training, we enable a student  and a teacher pipeline to achieve self-distillation. 

Formally, let $\mathcal{F}$ be the transformer encoder with its output  embedding dimension denoted by $D$, $\mathcal{P}$  the projector, $\mathcal{M}$   the masking operator, and $\mathcal{A}$  the aggregating operator. 
We start from explaining the student pipeline. 
An input image is randomly augmented to two distorted views: $x_1$ and $x_2$. 
Each view is subsequently divided into $HW$ non-overlapping patch tokens, denoted as $T_1= \left\{t_1^{(i)}\right\}_{i=1}^{HW}$ and $T_2=\left\{t_2^{(i)}\right\}_{i=1}^{HW}$, respectively. 
We forward $T_1$ into the encoder to obtain the logits $Z_1=\mathcal{F}(T_1) \in \mathbb{R}^{HW \times D}$, which are then fed into the classifier  for classification, and also the decoder for segmentation, following the standard image classification and segmentation setup.
To reinforce features, we divide $T_2$ into uncertain and confident tokens and mask the uncertain ones with learnable parameters, for which the uncertain token selection and masking approaches will be explained later. 
The resulting masked view, denoted as $\hat{T}_2=\mathcal{M}(T_2)$, is also fed into the encoder to obtain $\hat{Z}_2=\mathcal{F}\left(\hat{T}_2\right)$. 
Embeddings of the unmasked confident tokens in $\hat{Z}_2$ are summarized into a class token by an aggregation module, denoted by  $\mathcal{A}\left(\hat{Z}_2\right) \in \mathbb{R}^{1\times D}$. 
This class token is  concatenated with $\hat{Z}_2$, and further projected and normalized to resemble probabilities distributions  in $\hat{P}_2 \in \mathbb{R}^{(1+HW)\times D} $, as
\begin{equation}
\label{eq_agg}
    \hat{P}_2 = \sigma\left( \mathcal{P}\left( \left[\mathcal{A}\left(\hat{Z}_2\right);\hat{Z}_2 \right] \right) \right),
\end{equation}
where $\sigma$ is the row-wise softmax function, and $[;]$ the concatenation. We will explain the aggregation design  later.

The teacher shares the same architecture as the student's encoder and projector, and has a similar pipeline described by Eq. (\ref{eq_agg}), except it takes the unmasked inputs $T_1$ and $T_2$, and returns two distributions $P_1$ and $P_2$ for the two views, respectively. 
The student output  $ \hat{P}_2 $ and the teacher outputs $P_1$ and $P_2$ are used  for feature reinforcement training.

\subsection{Uncertain Patch Token Selection}
We select uncertain  patch tokens under the guidance of  semantic-aware CAMs, generated using the  logits  computed earlier with the first  view, i.e., $Z_1=\mathcal{F}(T_1)$. 
We  linearly project  $Z_1$ using the weights  $W \in \mathbb{R}^{C \times D}$ of the  classifier for image classification, where $C$ is the class number, and then normalize it by the $\operatorname{ReLU}$ function and $\operatorname{min-max}$ normalization. The  normalized CAM, denoted as $M_c \in \mathbb{R}^{HW \times C} (0 \leq M_C \leq 1)$, is defined by
\begin{equation}
    M_c := \operatorname{min-max}\left(\operatorname{ReLU}\left(ZW^{\top}\right)\right).
\end{equation}
It encodes the semantic uncertainty for each patch driven by CAM scores $ZW^{\top}$.

Next, we identify the uncertain regions based on the normalized CAM  and mask the uncertain patches, following an adaptive masking strategy. 
Features in non-reliable regions are considered as uncertain features. 
However, some reliable regions can be wrongly labeled, and their corresponding features  can also be uncertain. 
To remedy this, we propose an adaptive masking strategy, resulting in a soft masking vector $M_s \in \mathbb{R}^{HW}$ with each element given as
\begin{equation}
    M_s^{(i)}=\left\{
    \begin{array}{lr}
    u_i+1, & \text{if } \beta_l < \max \left(M_c^{(i,:)}\right) < \beta_h, \\
    u_i, & \text{otherwise},
    \end{array}\right.
\end{equation}
where $u_i \sim \text{U}(0,1)$ draws from a standard uniform distribution and enables a stochastic selection process. 
The above use of two background thresholds  $0 < \beta_l < \beta_h < 1$ for dividing patches into reliable and non-reliable ones is inspired by~\citet{zhang2020reliability} and~\citet{ru2022learning}, which suggests an intermediate score to be a sign of uncertainty.
As a result, elements in $M_s$ with larger values suggest uncertain patches. 

We use a masking ratio  $0<r<1$ to control the amount of selected uncertain patches, and defines the following binary selection mask $M_b \in \mathbb{R}^{HW}$ with each element   as
\begin{equation}
\label{Eq_Mb}
    M_b^{(i)}=\left\{
    \begin{array}{lr}
    1, & \text{if } i \in \argtopk(M_s), k:= \lfloor HW * r \rfloor, \\
    0, & \text{otherwise},
    \end{array}\right. 
\end{equation}
where  $\lfloor \cdot \rfloor$ denotes the floor function. 
The  selected uncertain patches, flagged by 1 in $M_b$, correspond to those top-$k$ large-valued elements in $M_s$. 
Our masking strategy is designed to relax the hard foreground-background thresholds by the masking ratio $r$. When more patches are flagged as uncertain by  $\beta_l < \max \left(M_c^{(i,:)}\right) < \beta_h$, the selection is randomly conducted within them through $u_i$. When less uncertain patches are flagged, part of confident patches are also selected. 
The original features of the selected tokens to mask are replaced by learnable parameters with the same feature dimension.

\begin{figure}[t]
    \centering
    \includegraphics[width=0.9\linewidth]{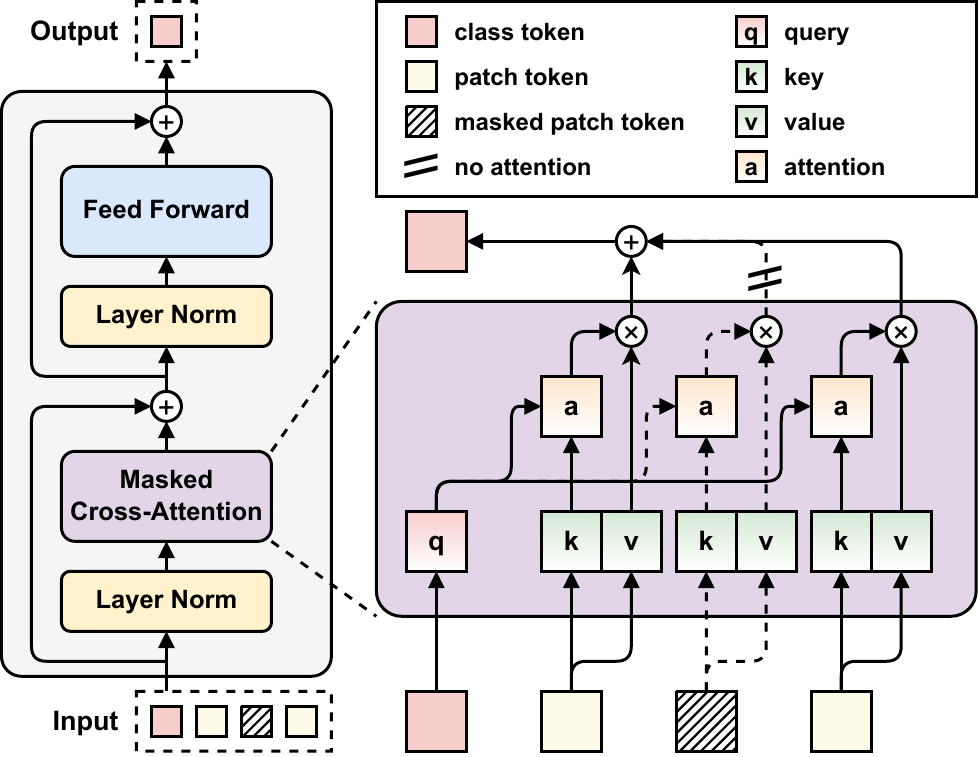}
    \caption{Illustration of the aggregation module. This module is composed of several aggregation blocks, where each block alternates in turn a cross-attention layer and a feed-forward layer. The cross-attention layer computes attention between a class token and a sequence of unmasked patch tokens.}
    \label{fig:aggregation}
\end{figure}

\subsection{Certain Feature Aggregation}
We design an attentive aggregation module to compress the embeddings of a sequence of
$N=HW$ patch tokens, stored in $\hat{Z} \in \mathbb{R}^{N \times D}$, into one class token embedding $\bar{Z} \in \mathbb{R}^{1 \times D}$.
 As shown in Figure~\ref{fig:aggregation}, this module contains several aggregation blocks, where each block contains a masked cross-attention (MCA) layer and a feed-forward (FF) layer, given as
\begin{equation}
    \begin{aligned}
    \bar{Z}^{(l)}_{(o)} & =\bar{Z}^{(l)}+\text{MCA}\left(\eta\left(\left[\bar{Z}^{(l)};\hat{Z}^{(l)}\right]\right)\right), \\
    \bar{Z}^{(l+1)} & =\bar{Z}^{(l)}_{(o)}+\text{FF}\left(\eta\left(\bar{Z}^{(l)}_{(o)}\right)\right),
    \end{aligned}
\end{equation}
where $l$ denotes the layer index and $\eta$ is the LayerNorm~\cite{ba2016layer}.

MCA is analogous to self-attention~\cite{vaswani2017attention}, except that it computes attention between the class token and a set of unmasked patch tokens. We parameterize MCA with projection weights $W_Q,W_K,W_V,W_O \in \mathbb{R}^{D \times D}$, and calculate the queries $Q \in \mathbb{R}^{1 \times D}$, 
keys $K \in \mathbb{R}^{N \times D}$ and values $V \in \mathbb{R}^{N \times D}$ by projection, so that
\begin{equation}
    Q = \eta\left( \bar{Z} \right) W_Q^{\top},
    K = \eta\left( \hat{Z} \right) W_K^{\top},
    V = \eta\left( \hat{Z} \right) W_V^{\top}.
\end{equation}
Note that queries are derived from the class token, while keys and values are calculated on patch tokens. The masked cross-attention $A \in \mathbb{R}^{1 \times N}$ is then formulated as 
\begin{equation}
    A = \sigma \left( \frac{\left(1-M_b\right) \left( QK^{\top} \right)}{\sqrt{D}} \right).
\end{equation}
The output of MCA is computed as a weighted sum of values, i.e., $\left(AV\right)W_O^{\top}$.

\subsection{Feature Self-reinforcement}
We adopt self-distillation~\cite{caron2021emerging,zhou2021ibot,oquab2023dinov2} to improve the model training for feature reinforcement. 
As explained earlier, given two distorted views of the same image, we compute one student output $\hat{P}_2$ and two teacher outputs $P_1$ and $P_2$, 
where their first element stores the aggregated token information, while the remaining the individual token content.
We propose a self-reinforcement loss $\mathcal{L}_{u}$ for the uncertain tokens, as the cross-entropy loss between each student's patch token and its corresponding teacher's patch token:
\begin{equation}
    \mathcal{L}_{u} = -\sum_{i=2}^{1+N} M_b^{(i)} P_2^{(i)} \log \hat{P}_2^{(i)},
\end{equation}
where $M_b$ is the mask in Eq. (\ref{Eq_Mb}) to help select masked patch tokens.
We also conduct  self-reinforcement  for the confident tokens, 
formulated as the cross-entropy loss on the two aggregated class tokens of the two views, given as
\begin{equation}
    \mathcal{L}_{c} = -P_1^{(1)} \log \hat{P}_2^{(1)}.
\end{equation}
Following a common practice, we adopt the multi-label soft margin loss $\mathcal{L}_{cls}$ for classification, the pixel-wise cross-entropy loss $\mathcal{L}_{seg}$ for segmentation, and the cosine similarity loss $\mathcal{L}_{aff}$ for affinity regularization. Denote the weighting factors as $\{\lambda_{i}\}_{i=1}^5$, the overall training objective is
\begin{equation}
\mathcal{L} = \lambda_{1} \mathcal{L}_{cls} + \lambda_{2} \mathcal{L}_{aff} + \lambda_{3} \mathcal{L}_{seg} 
    +\lambda_{4} \mathcal{L}_{u} + \lambda_{5} \mathcal{L}_{c}.
\end{equation}
It consolidates classification, segmentation and feature self-reinforcement within a single-stage framework. 

\section{Experiments}

\begin{table}[t]
    \centering
    \small
    \setlength{\tabcolsep}{1mm}
    \begin{tabular}{l|c|c|cc}
        \toprule
        Method & Sup. & Net. & Val & Test \\
        \midrule
        \multicolumn{5}{l}{\textbf{\textit{Multi-stage WSSS methods}}.} \\
        RIB~\cite{lee2021reducing}  & $\mathcal{I}+\mathcal{S}$ & RN101 & 70.2 & 70.0 \\
        EDAM~\cite{wu2021embedded}  & $\mathcal{I}+\mathcal{S}$ & RN101 & 70.9 & 70.6 \\
        EPS~\cite{lee2021railroad}  & $\mathcal{I}+\mathcal{S}$ & RN101 & 71.0 & 71.8 \\
        SANCE~\cite{li2022towards}  & $\mathcal{I}+\mathcal{S}$ & RN101 & 72.0 & 72.9 \\
        L2G~\cite{jiang2022l2g}     & $\mathcal{I}+\mathcal{S}$ & RN101 & 72.1 & 71.7 \\
        RCA~\cite{zhou2022regional} & $\mathcal{I}+\mathcal{S}$ & RN38  & 72.2 & 72.8 \\
        SEAM~\cite{wang2020self}        & $\mathcal{I}$ & RN38  & 64.5 & 65.7 \\
        BES~\cite{chen2020weakly}       & $\mathcal{I}$ & RN101 & 65.7 & 66.6 \\
        CPN~\cite{zhang2021complementary}&$\mathcal{I}$ & RN38  & 67.8 & 68.5 \\
        CDA~\cite{su2021context}        & $\mathcal{I}$ & RN38  & 66.1 & 66.8 \\
        ReCAM~\cite{chen2022class}      & $\mathcal{I}$ & RN101 & 68.5 & 68.4 \\
        URN~\cite{li2022uncertainty}    & $\mathcal{I}$ & RN101 & 69.5 & 69.7 \\
        ESOL~\cite{li2022expansion}     & $\mathcal{I}$ & RN101 & 69.9 & 69.3 \\
        $\dagger$ViT-PCM~\cite{rossetti2022max}         & $\mathcal{I}$ & RN101 & 70.3 & 70.9 \\
        $\dagger$MCTformer~\cite{xu2022multi}           & $\mathcal{I}$ & RN38  & 71.9 & 71.6 \\
        $\dagger$OCR~\cite{cheng2023out}                & $\mathcal{I}$ & RN38  & 72.7 & 72.0 \\
        $\dagger$BECO~\cite{rong2023boundary}           & $\mathcal{I}$ & MiT-B2& 73.7 & 73.5 \\
        $\dagger$MCTformer+~\cite{xu2023mctformerplus}  & $\mathcal{I}$ & RN38  & 74.0 & 73.6 \\
        \midrule
        \multicolumn{5}{l}{\textbf{\textit{Single-stage WSSS methods}}.} \\
        RRM~\cite{zhang2020reliability}     & $\mathcal{I}$ & RN38  & 62.6 & 62.9 \\
        1Stage~\cite{araslanov2020single}   & $\mathcal{I}$ & RN38  & 62.7 & 64.3 \\
        $\dagger$AFA~\cite{ru2022learning}  & $\mathcal{I}$ & MiT-B1& 66.0 & 66.3 \\
        $\dagger$ToCo~\cite{ru2023token}    & $\mathcal{I}$ & ViT-B & 71.1 & 72.2 \\
        $\dagger$\textbf{Ours}        & $\mathcal{I}$ & ViT-B & \textbf{75.7} & \textbf{75.0} \\
        \bottomrule
    \end{tabular}
    \caption{Performance comparison of semantic segmentation on PASCAL VOC 2012 in terms of mIoU(\%). Sup. denotes the supervision type. $\mathcal{I}$: image-level class labels. $\mathcal{S}$: off-the-shelf saliency maps. Net. denotes the segmentation network for multi-stage methods or the backbone for single-stage methods. RN38: Wide ResNet38~\cite{wu2019wider}, RN101: ResNet101~\cite{he2016deep}, MiT: Mix Transformer~\cite{xie2021segformer}. $\dagger$ flags transformer based classification network or  backbone.}
    \label{tab:sota_voc}
\end{table}

\begin{table}[t]
    \centering
    \small
    \begin{tabular*}{\linewidth}{@{\extracolsep{\fill}}l|c|c|c}
        \toprule
        Method & Sup. & Net. & Val \\
        \midrule
        \multicolumn{4}{l}{\textbf{\textit{Multi-stage WSSS methods}}.} \\
        EPS~\cite{lee2021railroad}      & $\mathcal{I}+\mathcal{S}$ & RN101 & 35.7 \\
        RIB~\cite{lee2021reducing}      & $\mathcal{I}+\mathcal{S}$ & RN101 & 43.8 \\
        L2G~\cite{jiang2022l2g}         & $\mathcal{I}+\mathcal{S}$ & RN101 & 44.2 \\
        CDA~\cite{su2021context}        & $\mathcal{I}$ & RN38  & 33.2 \\
        URN~\cite{li2022uncertainty}    & $\mathcal{I}$ & RN101 & 40.7 \\
        ESOL~\cite{li2022expansion}     & $\mathcal{I}$ & RN101 & 42.6 \\
        $\dagger$MCTformer~\cite{xu2022multi}           & $\mathcal{I}$ & RN38  & 42.0 \\
        $\dagger$ViT-PCM~\cite{rossetti2022max}         & $\mathcal{I}$ & RN101 & 45.0 \\
        $\dagger$OCR~\cite{cheng2023out}                & $\mathcal{I}$ & RN38  & 42.5 \\
        BECO~\cite{rong2023boundary}                    & $\mathcal{I}$ & RN101 & 45.1 \\
        $\dagger$MCTformer+~\cite{xu2023mctformerplus}  & $\mathcal{I}$ & RN38  & 45.2 \\
        \midrule
        \multicolumn{4}{l}{\textbf{\textit{Single-stage WSSS methods}}.} \\
        $\dagger$AFA~\cite{ru2022learning}  & $\mathcal{I}$ & MiT-B1    & 38.9 \\
        $\dagger$ToCo~\cite{ru2023token}    & $\mathcal{I}$ & ViT-B     & 42.3 \\
        $\dagger$\textbf{Ours}              & $\mathcal{I}$ & ViT-B     & \textbf{45.4} \\
        \bottomrule
    \end{tabular*}
    \caption{Performance comparison of semantic segmentation on MS COCO 2014 in terms of mIoU(\%). We use the same notations as in Table~\ref{tab:sota_voc}.}
    \label{tab:sota_coco}
\end{table}

\subsection{Experimental Settings}

\subsubsection{Datasets}
We evaluate our method on two benchmarks: PASCAL VOC 2012~\cite{everingham2010pascal} and MS COCO 2014~\cite{lin2014microsoft}. PASCAL VOC contains 20 object classes and one background class. Following the common practice of previous works~\cite{zhang2020reliability,araslanov2020single,ru2022learning,ru2023token}, it is augmented with data from the SBD dataset~\cite{hariharan2011semantic}, resulting in $10,582$, $1,449$ and $1,456$ images for training, validation and testing, respectively. MS COCO contains 80 object classes and one background class. It has $82,081$ images for training and $40,137$ images for validation. Note that we only adopt image-level labels during the training phase. We report mean Intersection-over-Union (mIoU) as the evaluation metric.

\subsubsection{Implementation Details}
We adopt ViT-B~\cite{dosovitskiy2020image} pretrained on ImageNet~\cite{deng2009imagenet} as the transformer encoder. The convolutional decoder refers to DeepLab-LargeFOV~\cite{chen2017deeplab}. We use two aggregation blocks in the aggregation module. The projector comprises a 3-layer perceptron and a weight-normalized fully connected layer~\cite{caron2021emerging}. Parameters in the aggregation module and the projector are randomly initialized. We use a light data augmentation: random resized cropping to $448 \times 448$ with the scale $[0.32, 1.0]$ and the ratio $[3/4,4/3]$, random horizontal flipping, and random color jittering. The student network is optimized with AdamW~\cite{loshchilov2017decoupled}. The base learning rate is warmed up to $6e-5$ at the first 1,500 iterations and decayed with a cosine schedule. The weighting factors $(\lambda_1, \lambda_2, \lambda_3, \lambda_4, \lambda_5)$ are set to $(1.0,0.2,0.1,0.1,0.1)$. The teacher network requires no gradient and is updated with the EMA momentum. Experimentally, we find that synchronizing the teacher encoder with the student (i.e., momentum is $0.0$) works better. The momentum for the teacher projector is $0.996$ and increases to $1.0$ with a cosine schedule during training. We embrace the centering and sharpening technique suggested in~\cite{caron2021emerging} to avoid collapsed solutions. The masking ratio $r$ is $0.4$ for adaptive uncertain feature selection. The background scores $(\beta_l, \beta_h)$ introduced to determine uncertain regions are $(0.2,0.7)$. Training iterations are 20,000 for PASCAL VOC 2012 and 80,000 for MS COCO 2014. We use multi-scale testing and dense CRF~\cite{chen2014semantic} at test time following~\cite{ru2022learning,ru2023token}. 

\subsection{Comparison with State-of-the-arts}

\paragraph{PASCAL VOC 2012}
Table~\ref{tab:sota_voc} shows comparison results of our proposed Feature Self-Reinforcement (FSR) with other state-of-the-art methods on PASCAL VOC 2012. As can be seen from this table, FSR significantly outperforms other single-stage approaches, achieving $75.7\%$ and $75.0\%$ mIoU on the validation and test sets, respectively. It is noticeable that our method achieves even higher mIoU than several sophisticated multi-stage methods, e.g., FSR surpasses BECO~\cite{rong2023boundary} by margins of $2.0\%$ and $1.5\%$. Compared with multi-stage methods using both image-level labels and off-the-shelf saliency maps, e.g., L2G~\cite{jiang2022l2g} and RCA~\cite{zhou2022regional}, our method still achieves superior performance. We assume although saliency maps are effective in providing additional background clues, our method can strengthen both confident regions (mostly the main body of objects or the background) and uncertain regions (mostly object boundaries), so that semantically distinct objects can be better differentiated. Moreover, it shows that recent methods with transformer-based networks (with $\dagger$) generally outperform those with convolutional networks (without $\dagger)$. Nevertheless, due to the difficulty of end-to-end optimization, single-stage transformer-based methods (e.g., ToCo reports $71.1\%$ and $72.2\%$) can only achieve comparable performance with multi-stage ones (e.g., BECO reports $73.7\%$ and $73.5\%$). Our method proves the efficacy of transformer-based single-stage training by attaining even better results. 

\paragraph{MS COCO 2014}
Table~\ref{tab:sota_coco} gives comparison results of semantic segmentation on a more challenging benchmark MS COCO 2014. We achieve $45.5\%$ mIoU on the validation set, which outperforms previous single-stage solutions and is slightly better than multi-stage MCTformer+~\cite{xu2023mctformerplus} by $0.2\%$. This further demonstrates the superiority of our proposed method. 

\begin{figure}[t]
    \centering
    \includegraphics[width=1.0\linewidth]{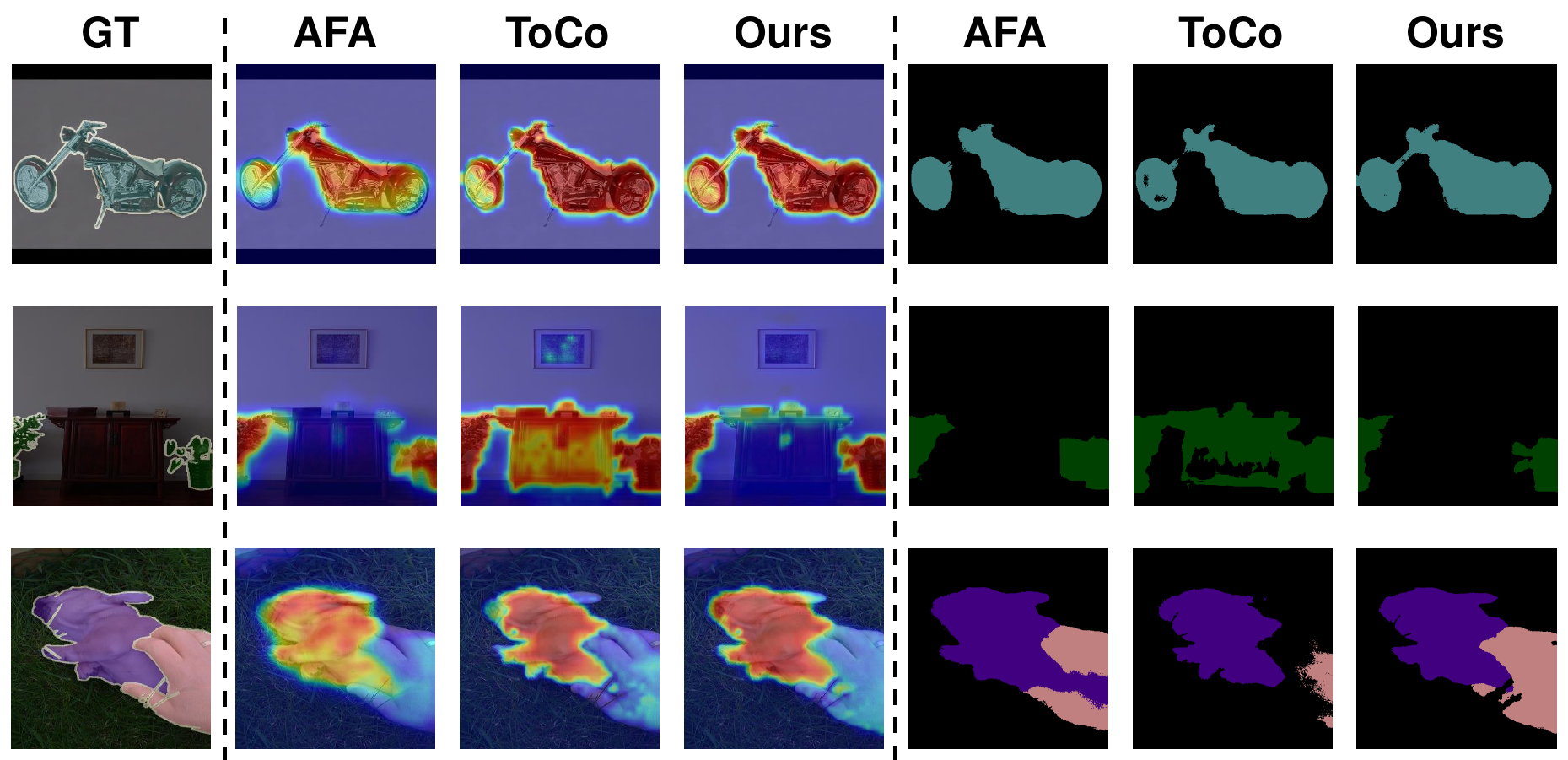}
    \caption{Visualization results of CAMs and predicted segmentation labels with SOTA single-stage frameworks (i.e., AFA and ToCo). (left) Ground truth. (middle) Comparison results of CAMs. (right) Comparison results of predicted segmentation labels. }
    \label{fig:result}
\end{figure}

\paragraph{Visualization Results}
In Figure~\ref{fig:result}, we visualize CAMs derived from the classifier and semantic segmentation labels predicted by the decoder of three single-stage methods, i.e., AFA~\cite{ru2022learning}, ToCo~\cite{ru2023token} and our proposed FSR. Compared with AFA, ToCo and FSR can generate more integral and deterministic CAMs. For instance, the wheels of ``motorbike'' are mildly activated by AFA while strongly confirmed by ToCo and FSR. This proves the effectiveness of FSR for uncertain features. However, AFA only focuses on boosting uncertain features, whereas our method enhances both uncertain and certain ones. For instance, AFA mistakes ``drawer'' as ``chair'', while FSR successfully recognizes the different semantics. This shows the importance of FSR for seemingly certain features. 

\subsection{Ablation Studies}
In this section, we present extensive ablation studies to verify the effectiveness of our proposed FSR. We report segmentation performance of pseudo labels (Pseu.) derived from CAMs as well as predicted labels (Pred.) generated by the decoder. All results are evaluated on PASCAL VOC 2012 val set. Dense CRF is not applied with ablations. 

\begin{table}[tbp]
    \centering
    \small
    \resizebox{0.95\linewidth}{!}{
    \begin{tabular}{c|c|c|ccccc}
        \toprule
                   & Edge      & CAM       & \multicolumn{5}{c}{CAM} \\
        mask ratio & (strict)  & (strict)  & 0.1 & 0.2 & 0.3 & 0.4 & 0.5  \\
        \midrule
        \multicolumn{8}{l}{\textit{Pseu. label results (\%)}} \\
        random    & -    & -    & 73.1          & 73.6          & \textbf{74.1} & 74.2          & 73.2          \\
        uncertain & 73.3 & 74.0 & \textbf{74.1} & \textbf{74.2} & 73.9          & \textbf{74.4} & \textbf{73.7} \\
        \midrule
        \multicolumn{8}{l}{\textit{Pred. label results (\%)}} \\
        random    & -     & -   & 71.7          & 72.3          & 71.3          & 72.3          & 71.2          \\
        uncertain & 71.6 & 71.8 & \textbf{72.2} & \textbf{72.3} & \textbf{72.0} & \textbf{72.5} & \textbf{72.1} \\
        \bottomrule
    \end{tabular}}
    \caption{Ablation results of uncertain feature selection methods. ``random'' means random masking, ``uncertain'' means our adaptive masking strategy that gives priority to masking uncertain regions. }
    \label{tab:uncertain_feature_selection}
\end{table}

\begin{table}[tbp]
    \centering
    \small
    \begin{tabular}{c|cc|cc}
    \toprule
    Masking & unc.FSR & cer.FSR & Pseu. (\%) & Pred. (\%) \\
    \midrule
    - & & & 71.1 & 67.9 \\
    \midrule
    \multirow{5}{*}{CAM}
    & \checkmark &                  & 74.4$_{\textbf{\textcolor{red}{+3.3}}}$ & 72.5$_{\textbf{\textcolor{red}{+4.6}}}$ \\
    &            & \checkmark (GAP) & 72.3$_{\textbf{\textcolor{red}{+1.2}}}$ & 70.9$_{\textbf{\textcolor{red}{+3.0}}}$ \\
    &            & \checkmark (GMP) & 71.8$_{\textbf{\textcolor{red}{+0.7}}}$ & 70.0$_{\textbf{\textcolor{red}{+2.1}}}$ \\
    &            & \checkmark (MCA) & 75.2${_\textbf{\textcolor{red}{+4.1}}}$ & 73.3$_{\textbf{\textcolor{red}{+5.4}}}$ \\
    & \checkmark & \checkmark (MCA) & \textbf{75.7}$_{\textbf{\textcolor{red}{+4.6}}}$ & \textbf{73.6}$_{\textbf{\textcolor{red}{+5.7}}}$ \\
    \bottomrule
    \end{tabular}
    \caption{Ablation results of unc.FSR and cer.FSR. ``GAP'', ``GMP'', and ``MCA'' are  aggregation methods of cer.FSR.}
    \label{tab:ablations}
\end{table}

\subsubsection{Analysis of Uncertain Feature Selection}
In Table~\ref{tab:uncertain_feature_selection}, we compare two \emph{strict} selection methods for uncertain features: edge-based selection and CAM-based selection. For edge-based selection, we choose the conventional Canny edge detector to extract edges in an image and generate exact masks of these edges. Activation thresholds for CAM-based selection are $(0.2,0.7)$. CAM-based selection is marginally better than edge-based selection; the improvement continues when CAM-based selection is relaxed, i.e., uncertain features are not strictly but preferentially masked. Empirically, we find that $r=0.4$ gives the best result. In addition, uncertain feature masking achieves higher performance than random feature masking in most cases, showing it is important to reinforce uncertain features for semantics clarification. 

\begin{figure}[tbp]
    \centering
    \includegraphics[width=1.0\linewidth]{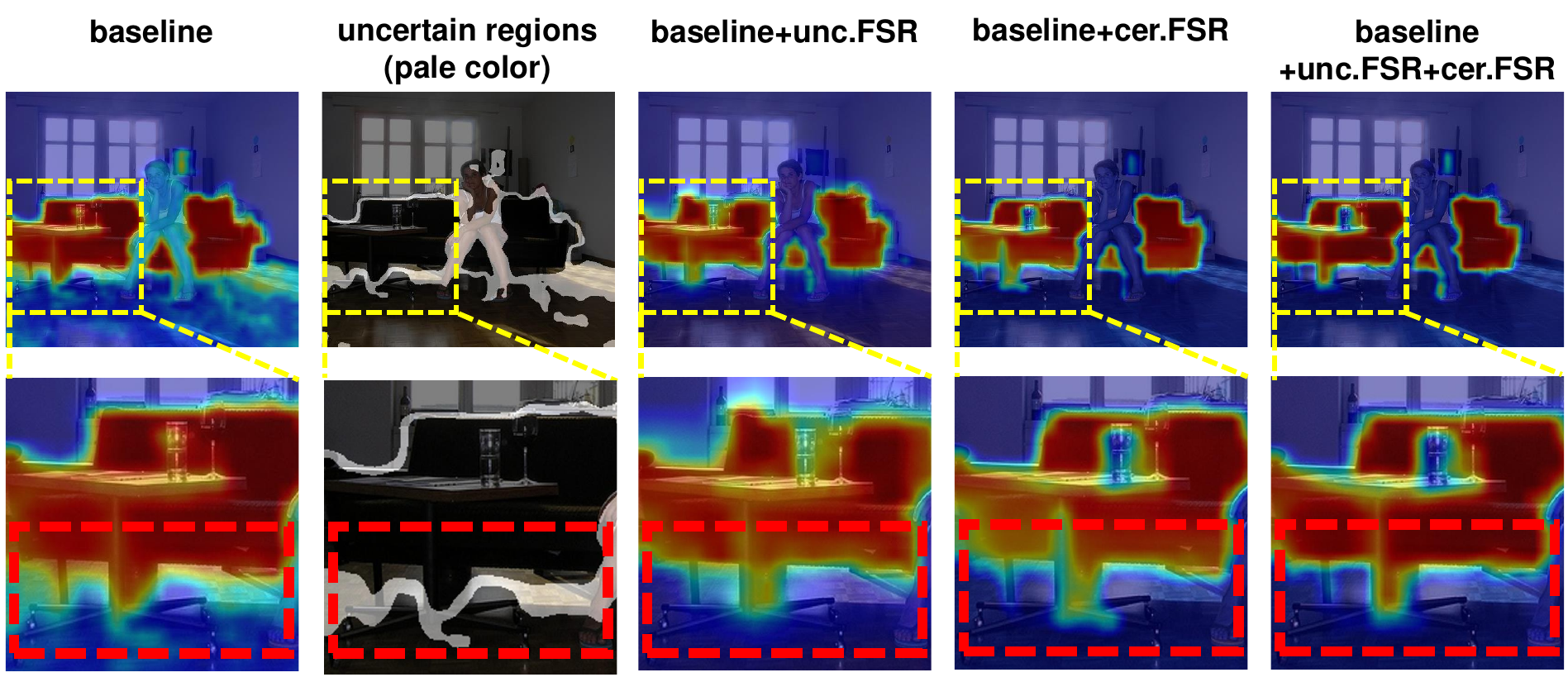}
    \caption{FSR optimizes the boundary regions (e.g., dashed red box) through the adaptive masking of regions characterized by uncertainty and the integration of unc.FSR and cer.FSR.}
    \label{fig:ablation_visualization}
\end{figure}

\subsubsection{Analysis of Feature Self-reinforcement}
Table~\ref{tab:ablations} shows the ablation results of FSR on uncertain regions (unc.FSR) and on certain regions (cer.FSR). The masking ratio is set to $0.4$ for comparison. It demonstrates the advancement of unc.FSR by achieving $74.4\%$ (compared to $71.1\%$) on pseudo labels and $72.5\%$ (compared to $67.9\%$) on predicted labels. This proves that reinforcing uncertain features, which mainly contain ambiguous object boundaries and misclassified categories, is fairly effective. When combining unc.FSR with cer.FSR, the quality of pseudo labels can be further improved, from $74.4\%$ to $75.7\%$; predicted labels directly supervised by pseudo labels are promoted as well, from $72.5\%$ to $73.6\%$. This indicates that reinforcing confident features is complementary to unc.FSR with enhanced global understanding. We showcase examples of our \textbf{FSR} strategy and its effect on object boundaries in Figure~\ref{fig:ablation_visualization}. 

\begin{figure}[tbp]
    \centering
    \includegraphics[width=1.0\linewidth]{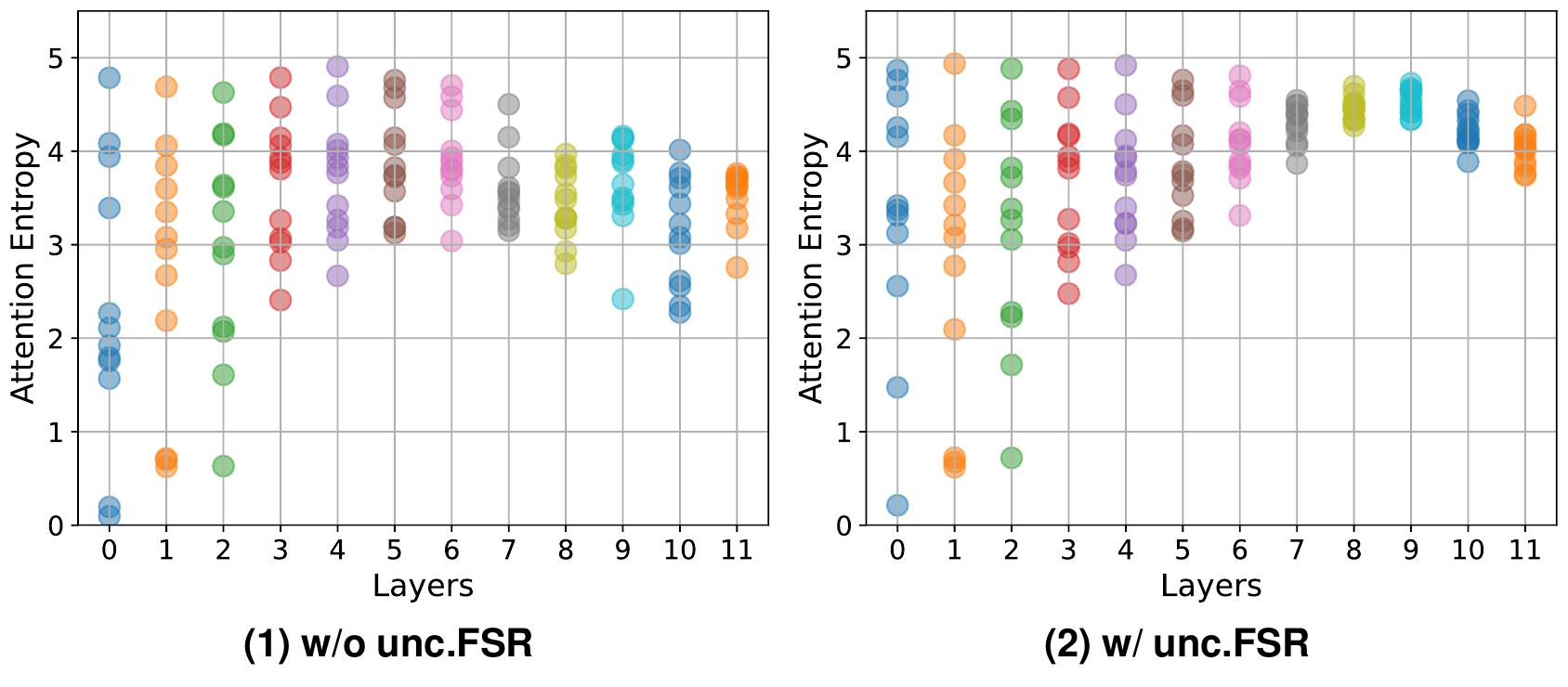}
    \caption{Average attention entropy of different attention heads (dots) across different layers.}
    \label{fig:attention_entropy}
\end{figure}


\paragraph{(a) Analysis of unc.FSR}
To gain a deep understanding of unc.FSR, we investigate the training process by analyzing the attention mechanism. Specifically, we compute average attention entropy~\cite{attanasio2022entropy} for each attention head across transformer layers. As shown in Figure~\ref{fig:attention_entropy}, the entropy at shallow layers (e.g., layer 0 to 6) holds similar without unc.FSR; however, it becomes higher and tighter at deep layers (e.g., layer 7 to 11) when unc.FSR is applied. A large entropy for a specific token indicates that a broad context contributes to this token, while a small entropy tells the opposite. From this point of view, we assume that unc.FSR benefits semantic segmentation by improving the degree of contextualization at deep layers. 


\begin{figure}[tbp]
    \centering
    \includegraphics[width=0.95\linewidth]{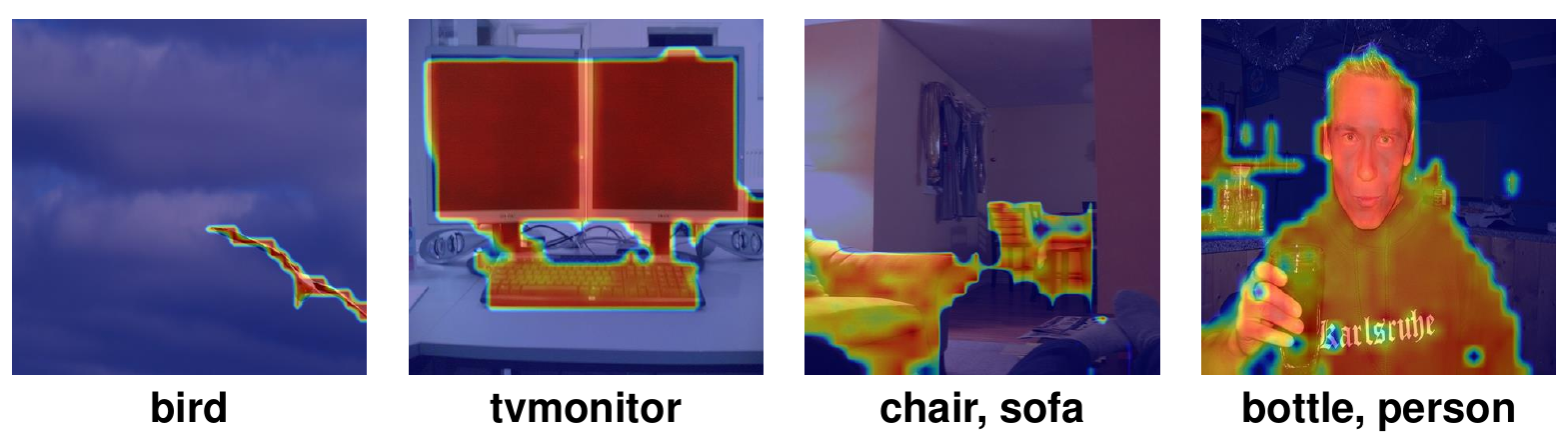}
    \caption{Class-to-patch attention maps derived from the aggregation module. Class labels are displayed below.}
    \label{fig:attention_maps}
\end{figure}

\paragraph{(b) Analysis of cer.FSR}
We compare our attentive aggregation of certain features (MCA) with two conventional methods: Global Average Pooling (GAP) and Global Maximum Pooling (GMP). GAP assigns an equal weight to each unmasked patch token, while GMP picks up the dominant one along each dimension. Table~\ref{tab:ablations} shows that GAP performs better than GMP, as GMP tends to intensify the most discriminative features, which may have an adverse effect in recognizing an integral object. It is noticeable that MCA outperforms GAP by a large margin, indicating an attentive weighting mechanism is superior to average weighting. We visualize class-to-patch attention maps in Figure~\ref{fig:attention_maps}, which illustrates that the class token can adaptively learn to pay attention to object regions. Note that the class token is not directly supervised by classification in our design; it interacts with unmasked patch tokens and learns to summarize effective information from them. 

\begin{table}[tbp]
    \centering
    \small
    \resizebox{\linewidth}{!}{
    \begin{tabular}{c|cccc}
        \toprule
        & Ours & +GaussianBlur & +Solarization & AutoAugment \\
        \midrule
        Pseu. (\%)   & 75.7          & \textbf{75.9} \textbf{\textcolor{teal}{$\pm$ 0.05}} & 75.3 \textbf{\textcolor{teal}{$\pm$ 0.12}} & 74.8 \textbf{\textcolor{teal}{$\pm$ 0.09}} \\
        Pred. (\%)   & \textbf{73.6} & \textbf{73.6} \textbf{\textcolor{teal}{$\pm$ 0.02}} & 73.2 \textbf{\textcolor{teal}{$\pm$ 0.06}} & 72.8 \textbf{\textcolor{teal}{$\pm$ 0.04}} \\
        \bottomrule
    \end{tabular}}
    \caption{10-trial experimental results of data augmentations. ``Ours'' is our default data augmentation setting. }
    \label{tab:augmentation}
\end{table}

\subsubsection{Data Augmentation}
We present comparison results with other data augmentations in Table~\ref{tab:augmentation}, which reveals that \textit{data augmentations have limited impacts on the performance}. For example, the performances display variations within the anticipated range when incorporating GaussianBlur or Solarization. Even when we substitute the data augmentation with the robust AutoAugmentation~\cite{cubuk2018autoaugment}, the results witness a slight decline as a strong augmentation may interfere with the segmentation objective. 

\section{Conclusion}
In this work, we propose to estimate boundaries with the guidance of semantic uncertainty identified by CAM. To achieve this, we design an activation-based masking strategy and seek to recover local information with self-distilled knowledge. We further introduce a self-distillation method to reinforce semantic consistency with another augmented view. We integrate our method into the single-stage WSSS framework and validate its effectiveness on PASCAL VOC 2012 and MS COCO 2014 benchmarks.

\section*{Acknowledgement}

This work was supported by the National Natural Science Foundation of China under Grant No. (62106235, 62202015, 62376206, 62003256), and in part by the Exploratory Research Project of Zhejiang Lab under Grant 2022PG0AN01.

{\small \bibliography{aaai24}}
\end{document}